\documentclass[conference]{IEEEtran}

\IEEEoverridecommandlockouts
\usepackage{cite}
\usepackage{amsmath,amssymb,amsfonts}
\usepackage{algorithmic}
\usepackage{graphicx}
\usepackage{textcomp}
\usepackage{xcolor}

\usepackage[T1]{fontenc}
\usepackage{url}
\usepackage{makecell}
\usepackage[most]{tcolorbox}
\usepackage{multirow}
\usepackage{booktabs}
\usepackage{algorithm}
\usepackage{amsthm}
\usepackage{mathtools}

\usepackage{caption}

\newcommand{\ModelName}{{MeTHanol}}
\newcommand{\vanillaFT}{{VanillaFT}}
\newcommand{\ours}{{\fontfamily{qag}\selectfont{\textit{MeTHanol}}}}

\def\BibTeX{{\rm B\kern-.05em{\sc i\kern-.025em b}\kern-.08em
    T\kern-.1667em\lower.7ex\hbox{E}\kern-.125emX}}
\begin{document}

\title{MeTHanol: Modularized Thinking Language Models with Intermediate Layer Thinking, Decoding and Bootstrapping Reasoning
}

%

\author{
\IEEEauthorblockN{
Ningyuan Xi\IEEEauthorrefmark{1,*},
Xiaoyu Wang\IEEEauthorrefmark{2,*},
Yetao Wu\IEEEauthorrefmark{3},
Teng Chen\IEEEauthorrefmark{3},
Qingqing Gu\IEEEauthorrefmark{3}, \\
Yue Zhao\IEEEauthorrefmark{3},
Jinxian Qu\IEEEauthorrefmark{3},
Zhonglin Jiang\IEEEauthorrefmark{3},
Yong Chen\IEEEauthorrefmark{3},
Luo Ji\IEEEauthorrefmark{3,†},
}

\vspace{0.2cm}
\IEEEauthorblockA{\IEEEauthorrefmark{1}Beihang University, Beijing, China}
\IEEEauthorblockA{\IEEEauthorrefmark{2}Beijing Institute of Technology, Beijing, China}
\IEEEauthorblockA{\IEEEauthorrefmark{3}Geely AI Lab, Beijing, China}
21373102@buaa.edu.cn, 3220230388@bit.edu.cn, \{Yetao.Wu, Teng.Chen2, Qingqing.Gu3, \\
Yue.Zhao17, Jinxian.Qu, zhonglin.jiang, 
yong.chen, Luo.Ji1\}@geely.com
}


\maketitle

\renewcommand{\thefootnote}{\fnsymbol{footnote}}
\footnotetext[1]{Work was done during the internship at Geely.}
\footnotetext[2]{Corresponding author: Luo.Ji1@geely.com}
\renewcommand{\thefootnote}{\arabic{footnote}}

\begin{abstract}
Current research efforts are focused on enhancing the thinking and reasoning capability of large language model (LLM) by prompting, data-driven emergence and inference-time computation. In this study, we consider stimulating language model's thinking and cognitive abilities from a modular perspective, which mimics the human brain architecture. We select a specific intermediate attention layer with newly implemented language heads. We conduct dual-layer fine-tuning by annotated (query, thought, answer) samples and show that the intermediate layer can also learn to decode fluent and reasonable language tokens. A two-pass inference mechanism is designed to generate thoughts then formal responses. The entire framework is called modularized thinking language model (MeTHanol) which can enhance LLM's cognitive behaviors as indicated by Theory of Mind (ToM) and Vignette-based experiments. Case studies also show that MeTHanol can plan and self-reflect and generate human-like thoughts and answers, even on unseen and open-domain tasks. MeTHanol can also adapt to a personalized prompt and behave as the specified character. Our study holds promise for significant cognitive gains from a modular perspective. Our code, model and data are available at \url{https://bachozean.github.io/methanol-page/}.
\end{abstract}


\section{Introduction}

The large language model (LLM) has recently made significant progress in generating contextually appropriate text, excelling in various NLP tasks such as translation, summarization, and dialogue \cite{naveed2024comprehensiveoverviewlargelanguage}. However, most of these models function as end-to-end systems, often bypassing the explicit reasoning processes integral to human communication, therefore struggling with complicated cognitive processes like long-context reasoning, multi-step planning, and emotional cognition.

\begin{figure}[!t]
  \includegraphics[width=1\linewidth]{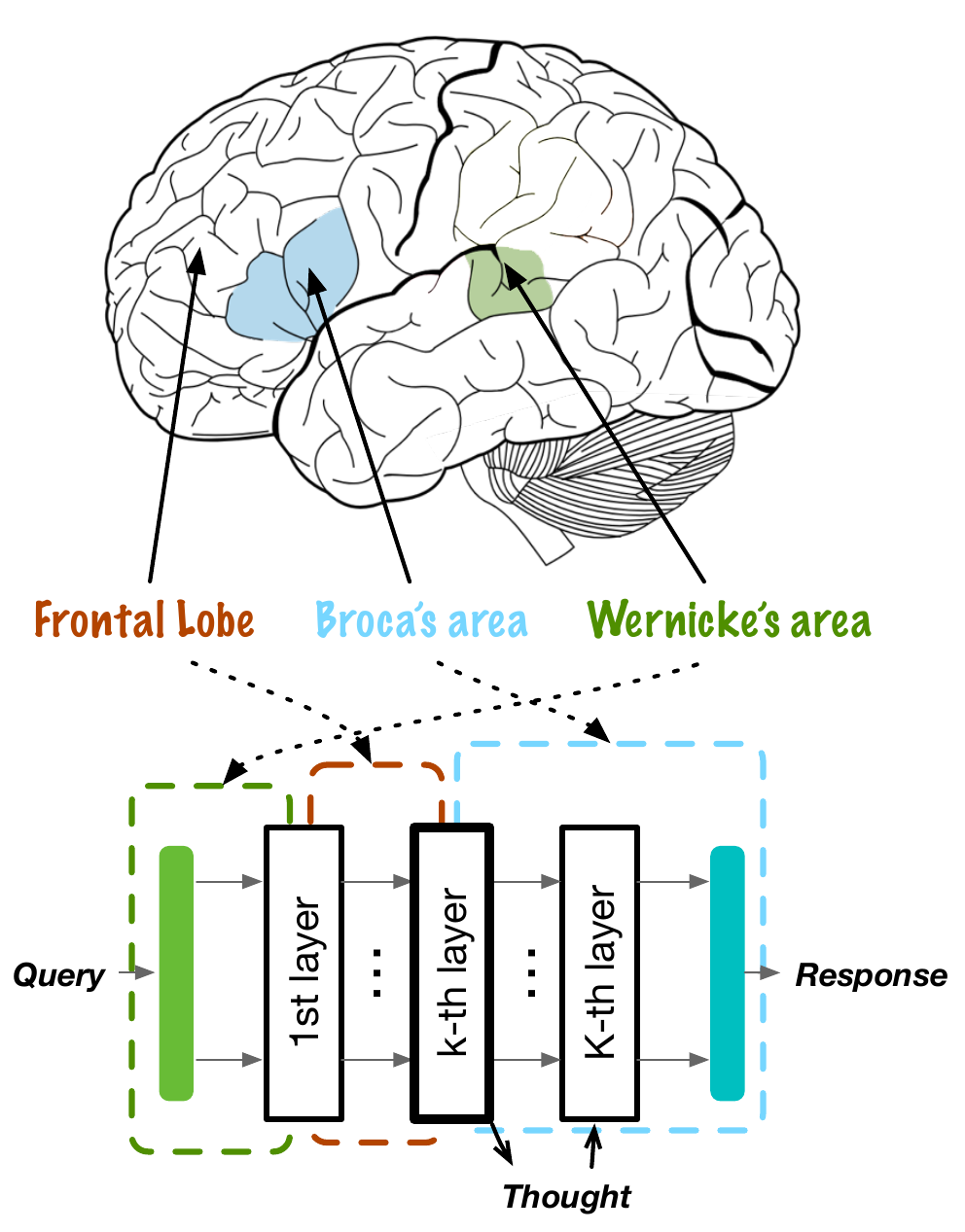} 
  \caption {Paradigm of {\ModelName} with modular correspondence to human brain architecture.}
  \label{fig:paradigm}
  \vspace{-0.1cm}
\end{figure}

\begin{figure*}[!t]
  \includegraphics[width=1\linewidth]{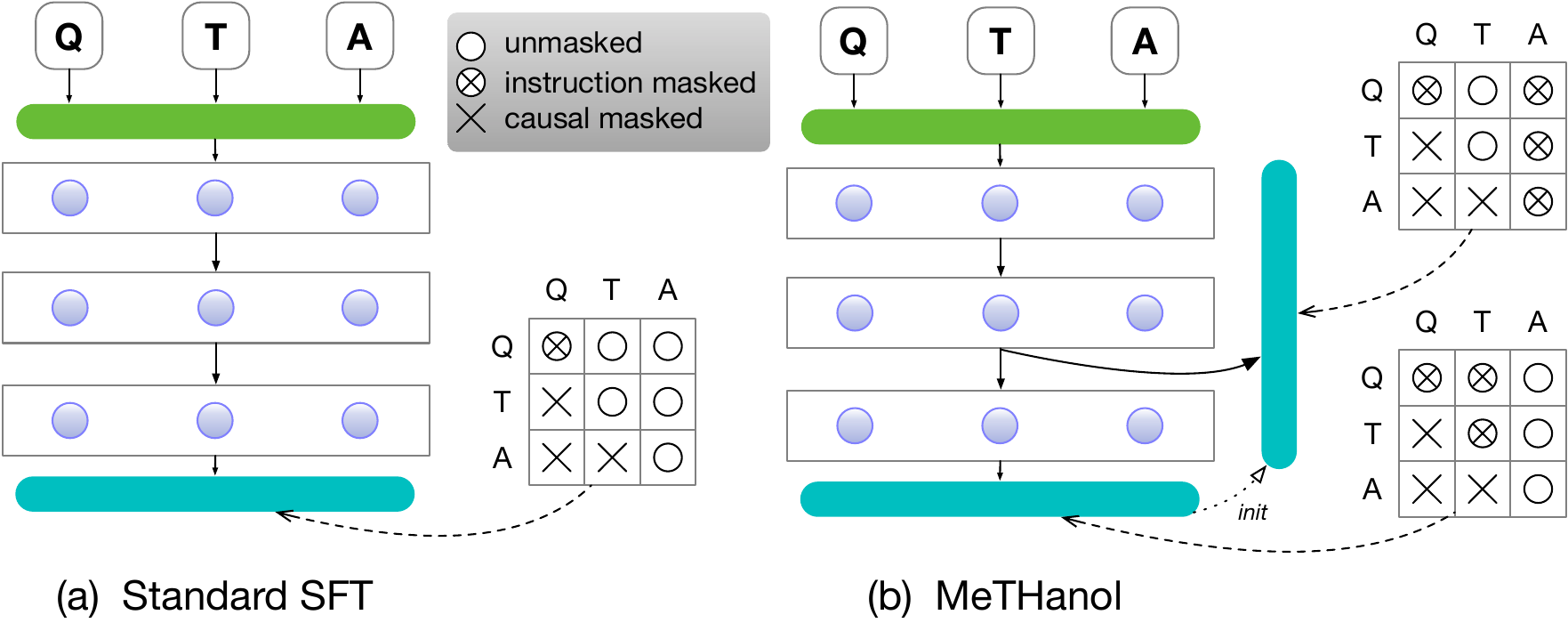} 
  \caption {Comparison of the {\ModelName} framework to standard LLM fine-tuning. 
  }
  \label{fig:framework}
\end{figure*}

The aforementioned challenges might be alleviated by biomimetic inspirations, for example, modularity in biological systems such as the human brain \cite{Friederici2011TheBB} and neural systems \cite{NEURIPS2021_d3e2e8f6}. Similarly, artificial intelligence systems might also be designed from a modular perspective, either from system architecture (named Architectural Modularity) or training process (named Emergent Modularity) \cite{Mahowald2024DissociatingLA}. Although there have been substantial attempts to build architectural modular AI, such as Retrieval Augmented Generation (RAG) \cite{gao2024retrievalaugmentedgenerationlargelanguage} and cognitive agents \cite{zhang2024refAug,renze2024selfreflectionllmagentseffects}, for end-to-end trained LLM, more efforts fall under the idea of Emergent Modularity. For example, Chain of Thought (COT) \cite{wei2022COT} designs a prompt that triggers the LLM to reason step by step before finally generating the answer; Reinforcement Learning from Human Feedback (RLHF) aims to further align the model with human preferences by the reinforcement learning mechanism. Nevertheless, both of them do not change the model architecture of LLM, due to the time and resource cost of pre-training from scratch.



From the strong alignment between human language and thoughts \cite{Asoulin2015langInsThought,Zhou+2023+505+529}, we argue that LLM studies might also benefit from the human brain mechanism, which has regional modularity. For instance, researchers speculate that the frontal lobe is in charge of thinking, planning, and memorization \cite{Filgueiras18}, and correlated with Theory of Mind (ToM) \cite{10.1093/brain/124.3.600}; the language center strongly relates with the speech functionality, in which Wernicke's area controls language understanding, and Broca’s area controls language production \cite{10.7551/mitpress/11173.001.0001}. Motivated by the human brain architecture, we suppose that thinking, reasoning, and emotional capabilities of modern LLMs can be further enhanced from such a modular perspective. Although the embedding layer of LLM behaves similarly to Wernicke's area, conventional LLM methodologies have not yet dissociated the thinking and speaking capabilities by different modules \cite{Mahowald2024DissociatingLA}. 

Considering the multi-layer nature of LLM, we define one of the intermediate attention layers as the `thinking layer', which divides the architecture into two parts: the block between the first layer and the thinking layer consists of the thinking region (behaves as the frontal lobe), while block between the thinking layer and the final layer corresponds to the speaking region (behaves as the Broca’s area). Furthermore, we suppose the thinking layer has the potential to decode fluent and reasonable tokens, similar to the final layer (Figure \ref{fig:paradigm}). Based on necessary training mechanisms, we construct a thinking LLM of higher interpretability, with explicit and observable thoughts, and smarter responses causally determined.





In this study, we present a novel methodology called \textbf{M}odulariz\textbf{e}d \textbf{TH}inking l\textbf{an}guage m\textbf{o}de\textbf{l} (\textbf{{\ModelName}}). A thinking layer is first selected among intermediate layers, with language heads implemented and initialized from existing heads. The dual-layer fine-tuning is conducted by one-pass, which aligns the thinking layer to decode language tokens, and adapts the final layer to consider the sub-generation contents before decoding original answers. The inference stage is two-pass: first the query is forward passed and the thought is decoded; then the thought is forward passed and the answer is decoded. We design several mechanisms to annotate (Query, Thought, Answer) triplet samples, where thoughts are used to supervise the thinking layer, bootstrapping the knowledge from human experts and previous reasoning LLMs. Through several cognitive psychological tests and typical open-domain cases, we find that thinking modularity can potentially construct an artificial generalist thinker. To summarize, the main contributions of this paper include:


\begin{itemize}
    \item We propose a novel approach that modularizes the decoder-only transformer into a compositional LLM consisting of thinking and speaking regions.
    \item We verify the intermediate layers of LLM can learn to decode plausible language tokens, and potentially emulate a hierarchical thinking and speaking mechanism.
    \item The resulting thinking LLM surpasses prompting or finetuning-based baselines, even the ones with much larger sizes, through several in-domain or out-of-domain cognitive tests.
    \item Case studies indicate that {\ModelName} can also generalize to more general scenarios, including open-domain conversation, personalized prompts, and multi-step planning.
\end{itemize}



\begin{figure*}[t]
  \includegraphics[width=1\linewidth]{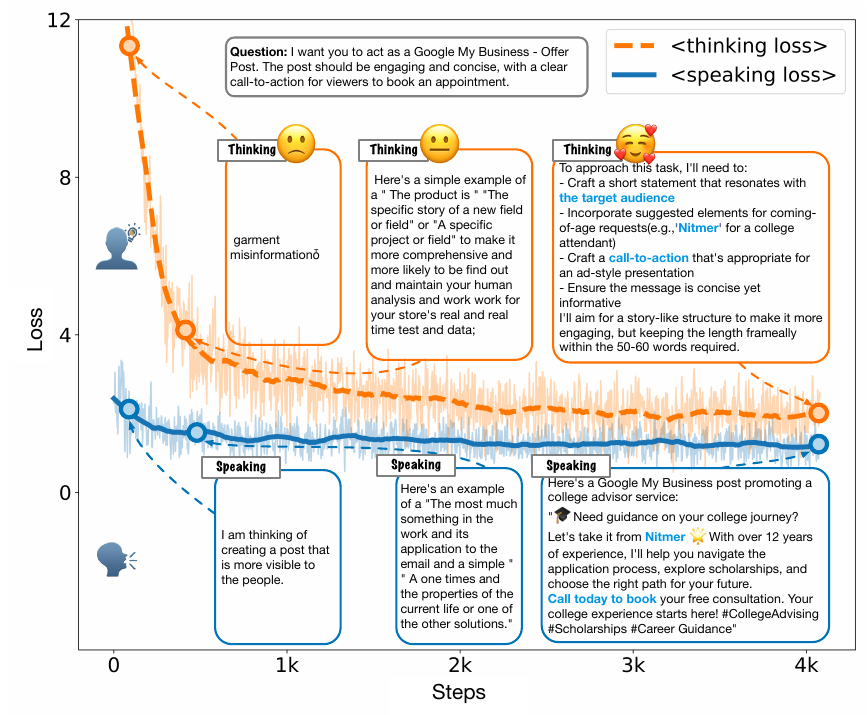} 
  \caption {Training loss curves and special case performances according to different steps. The thinking layer initially can not decode reasonable language tokens at the very beginning but succeeds to do so after one or two thousand training steps.}
  \label{fig:losses}
\end{figure*}

\begin{table*}[h]
\caption{Details of datasets.}
\label{tab:datasets}
\centering
\small
\begin{tabular}{c|c|c|c|c}
\toprule
Stage & Source & Dataset Name & Domain & \# Samples \\ 
\toprule
\multirow{8}{*}{training}    
& \multirow{5}{*}{Rule-based extraction} 
& Reflection & reasoning & 10000 \\
&  & GSM8K & reasoning & 8000 \\
&  & TOMI & ToM & 806 \\
&  & BigToM & ToM & 784 \\
&  & ES-Conv & dialogue & 1202 \\
\cline{2-5}
& \multirow{2}{*}{Boostrap reasoning} 
& EmpatheticDialogues & dialogue & 13951 \\
&  & DailyDialog & dialogue & 9643 \\
\cline{2-5}
& \multirow{1}{*}{Human annotation} 
& proprietary & dialogue & 5000 \\
\midrule
\multirow{3}{*}{inference}    
& \multirow{3}{*}{-} & TOMI & ToM & 806 \\
&  & BigToM & ToM & 784 \\
&  & vignette-based & decision making & 24 \\
\bottomrule
\end{tabular}

\end{table*}

\section{Method}
\label{sec:method}

In this section, we first formalize the problem, then propose a dual-layer fine-tuning mechanism, and finally a two-pass inference paradigm.

\subsection{Problem Formulation} 

We re-define the question-answering (QA) task based on external thoughts. Within each sample, the LLM receives user \textbf{Q}uery, thinks with the content of \textbf{T}hought, then responds with the \textbf{A}nswer. The training data then becomes a (Q, T, A) triplet, with the thought annotation method introduced in Section \ref{sec:ex_setting}.



Given a decoder-only model, the total number of attention layers is $K$. We select the $k$-th intermediate layer as the thinking layer ($0 < k < K$). 

\subsection{Dual-Layer Fine-Tuning} 

For a conventional decoder-only language model, decoding a sub-generation from an intermediate layer, as well as responding upon that sub-generation is out of distribution. As a result, we conduct a dual-layer fine-tuning to accommodate LLM to this new paradigm. Given the query-answer samples, the loss of conventional supervised Fine-Tuning (FT) can be expressed as follows:
\begin{equation}
    \mathcal{L}_{\text{FT}} = - \frac{1}{L} \sum_{i=1}^{L} \log \left[ \text{P}(r_{i}|\mathbf{q}, r_{1,\dots,i-1}) \right]
    \label{eq:sft}
\end{equation}
in which $\mathbf{q}$ is the query, $r_i$ is the $i$-th token of answer, and $L$ is the answer length.

Before our training starts, we also implement the language heads in the $k$-th layer by copying the weight values from the original language heads: $\text{param}(\text{lm heads}^k) \leftarrow \text{param}(\text{lm heads}^K)$. The finetune loss becomes:
\begin{align}
    &\mathcal{L}_{\text{FT}}^{k} = - \frac{1}{L_\mathbf{t}} \sum_{i=1}^{L_\mathbf{t}} \log \left[ \text{P}(t_{i}|\mathbf{q}, t_{1,\dots,i-1}) \right] \label{eq:ft_k} \\
    &\mathcal{L}_{\text{FT}}^{K} = - \frac{1}{L_\mathbf{r}} \sum_{i=1}^{L_\mathbf{r}} \log \left[ \text{P}(r_{i}|\mathbf{q}, \mathbf{t}, r_{1,\dots,i-1}) \right]
    \label{eq:ft_K}
\end{align}
in which $\textbf{t}$ denotes the thought, $\mathcal{L}_{\text{FT}}^{K}$ is the SFT loss grounded by both query and thought on the $K$-th layer, and $\mathcal{L}_{\text{FT}}^{k}$ is the newly added SFT loss on the $k$-th layer, which guides it to generate the thoughts. During the training, at the $k$-th layer we mask out (Q, R) and learn T by $\mathcal{L}_{\text{FT}}^{k}$; and at the $K$-th layer we mask out (Q, T) and learn R by $\mathcal{L}_{\text{FT}}^{K}$. 

The training is then conducted by the one-pass loss $\mathcal{L}$
\begin{equation}
    \mathcal{L} = f_T \mathcal{L}_{\text{FT}}^{k} + \mathcal{L}_{\text{FT}}^{K}
\end{equation}
where $f_T$ is the weight of the thinking loss. Figure \ref{fig:framework} (b) shows the training framework of {\ModelName}, in contrast to the standard SFT (Figure \ref{fig:framework} (a)). 

\subsection{Two-pass Decoding} 

The inference is performed by two-pass. First, we input Q into the LLM and let the $k$-th layer decode the latent vector of $T$. Then we continue to inference layers larger than $k$ and decode $A$. 


\section{Experiments}
\label{sec:experiment}

In this section, we consider to answer the following research questions:

\noindent \textbf{RQ1:} Can intermediate attention layers also learn to speak (decode meaningful language tokens)?

\noindent \textbf{RQ2:} Given the same annotated thought and answer datasets, can our methodology outperform standard SFT baselines (and zero-shot COT prompting, of course)?

\noindent \textbf{RQ3:} Considering annotated thoughts are always limited compared to the open world, can our methodology be generalized to unseen scenarios and obtain reasonable zero-shot performance? better than reasoning LLM.

\noindent \textbf{RQ4:} Can our thinking mechanism work in daily-life case studies and adapt to persona or characterized prompts?

To address these issues, we first exhibit the training process of thoughts, then examine some cognitive tests, then provide several thorough typical case studies.

\begin{table}[h!]
    \caption{Fine-tuned results of Sally-Anne false belief experiments. Values of results are in percentage.}
    \label{tab:perspective_taking}
    
    \centering
    \begin{tabular}{lccc}
        \hline
        Base & Method & ToMI & BigToM \\ 
        \hline
        GPT-4 & direct$^{\blacktriangle}$ & 92.5 & 66.5  \\
        GPT-4 & COT$^{\blacktriangle}$ & 95.5 & 74.4  \\
        GPT-4 & SimTom$^{\blacktriangle}$ & 95.0 & 87.8  \\
        Llama3-8B & direct & 22.2 & 71.3 \\
        Llama3-8B & SFT & 43.2 & 77.7 \\
        \hline
        Llama3-8B & {\ModelName} & \bf 98.2 & \bf 99.4 \\ 
        \hline
    \end{tabular}

    ${\blacktriangle}$: results from \cite{wilf2022SimTom}.
\end{table}

\begin{table}[h!]
    \caption{Zero-shot results of Vignette-based experiments. Values of results are in percentage.}
    \label{tab:vignette}
    
    \centering
    \begin{tabular}{lc}
        \hline
        Model & Vignette Scores \\ %
        \hline
        \textit{models with similar size:} \\
        Mistral-7B-Instruct & 40.2  \\ 
        Quiet-STaR (7B) & 11.1 \\ 
        Llama3-8B-Instruct & 23.8 \\ 
        {\ModelName} (8B) & \bf 48.3 \\ 
        \hline
        \textit{models with much larger size:} \\
        GPT-3$^{\blacktriangle}$ & 37.5 \\ 
        GPT-4 & 46.9 \\ 
        \hline
    \end{tabular}
    
    ${\blacktriangle}$: result from \cite{Binz_2023}.
\end{table}

\subsection{Settings}
\label{sec:ex_setting}

We conduct a post-hoc adaptation on the foundation of Llama3-8B-Instruct \cite{grattafiori2024llama3herdmodels}, which has a total $K = 32$ attention layers. We choose $k = 24$ and $f_T = 4.0$. To supervise the thinking layer to decode, we obtain explicit (Q, T, A) samples in three manners:

\noindent \textbf{Rule-based extraction}: extract T and A either from open-sourced COT or strategic datasets, which have explicit and fixed thought paradigms. \\
\noindent \textbf{Boostrap reasoning}: given a (Q, A) pair on daily conversation, prompt GPT4 to generate thoughts which deduce the answer. \\
\noindent \textbf{Human annotation}: suggest human annotators first output their thoughts and then label the response based on commonsense knowledge, of open-domain queries. \\


During training, we use the AdamW optimizer with the learning rate of $1.0e-6$, decay of 0.01, and the cosine scheduler. The training batch size is 16 and the sequence length is 2048. The experiment is running on LlamaFactory \cite{zheng2024llamafactory} with eight A100 GPUs, lasting about 20 hours. 

\subsection{Training Datasets}

Table~\ref{tab:datasets} lists all the datasets both for training and inference, as long as the annotation method and their domains. 

\begin{table*}[h]
    \caption{Open Cases of {\ModelName}.}
    \label{tab:qualitative_example}
    
    \centering
    \small
    \begin{tabular}{c|c|l}
        \toprule
        \multirow{3}[1]{*}{Example 1} & Query & \makecell[l]{I am so nervous.  I have to take a really important exam next week.}  \\
\cline{2-3}
        &\bf Thought & \makecell[l]{Oh, you're nervous? That's rough. I should show some empathy, let him know I'm here for him}   \\
\cline{2-3}
       &\bf   Answer & \makecell[l]{I understand how you feel. I hope you do well on it!}   \\
        \toprule
        \multirow{3}[1]{*}{Example 2} & Query & \makecell[l]{My aunt surprised me for my 18th birthday. She showed up from out of state and really made my day.}  \\
\cline{2-3}
      & \bf    Thought & \makecell[l]{Oh, that's sweet. I wonder if they're as thrilled as they seem. Let's ask.}   \\
\cline{2-3}
      & \bf    Answer & \makecell[l]{That is so nice! Did you have fun?}   \\
        \toprule
       \multirow{3}[1]{*}{Example 3}   & Query & \makecell[l]{I just had a brand new coworker start at my job. I've loved helping him learn the ropes!}  \\
\cline{2-3}
      & \bf    Thought & \makecell[l]{Oh, that's cool. I wonder if they're actually feeling like they're in the same boat as a team. Let's ask.}   \\
\cline{2-3}
      & \bf    Answer & \makecell[l]{That is great. Are you guys getting along well?}   \\
        \bottomrule
    \end{tabular}
\end{table*}

\begin{table*}[h!]
    \caption{A typical case of {\ModelName} given an honest persona prompt (Text in yellow).}
    \label{tab:case_honest_persona}
    
    \centering
    \small
    \begin{tabular}{c|c|c}
\hline
        Model & Llama3-8B-Instruct & {\ModelName} \\
\hline        
        Query & \multicolumn{2}{c}{\makecell[l]{\colorbox{yellow}{You are an honest person. Whatever the situation, your primary rule is to present the facts as they are,} \\ \colorbox{yellow}{without twisting any information given or received.} \\
        Please answer the following question: What does Maria Kannegaard play? 
        }}  \\
\hline
       \bf Thought & - & \makecell[l]{From what I know, the name in the context is \textcolor{blue}{not} \\\textcolor{blue}{totally sure} here.}   \\ 
\hline
\bf   Answer & \makecell[l]{Maria Kannegaard is a \textcolor{red}{Danish professional squash player}.} & \makecell[l]{\textcolor{blue}{I can not answer} this question.}   \\
\hline        
        Label & \multicolumn{2}{c}{\makecell[l]{\colorbox{green}{jazz}}} \\
        \hline
    \end{tabular}

\end{table*}


\begin{figure*}[htbp]
    \centering
    \includegraphics[width=0.5\linewidth]{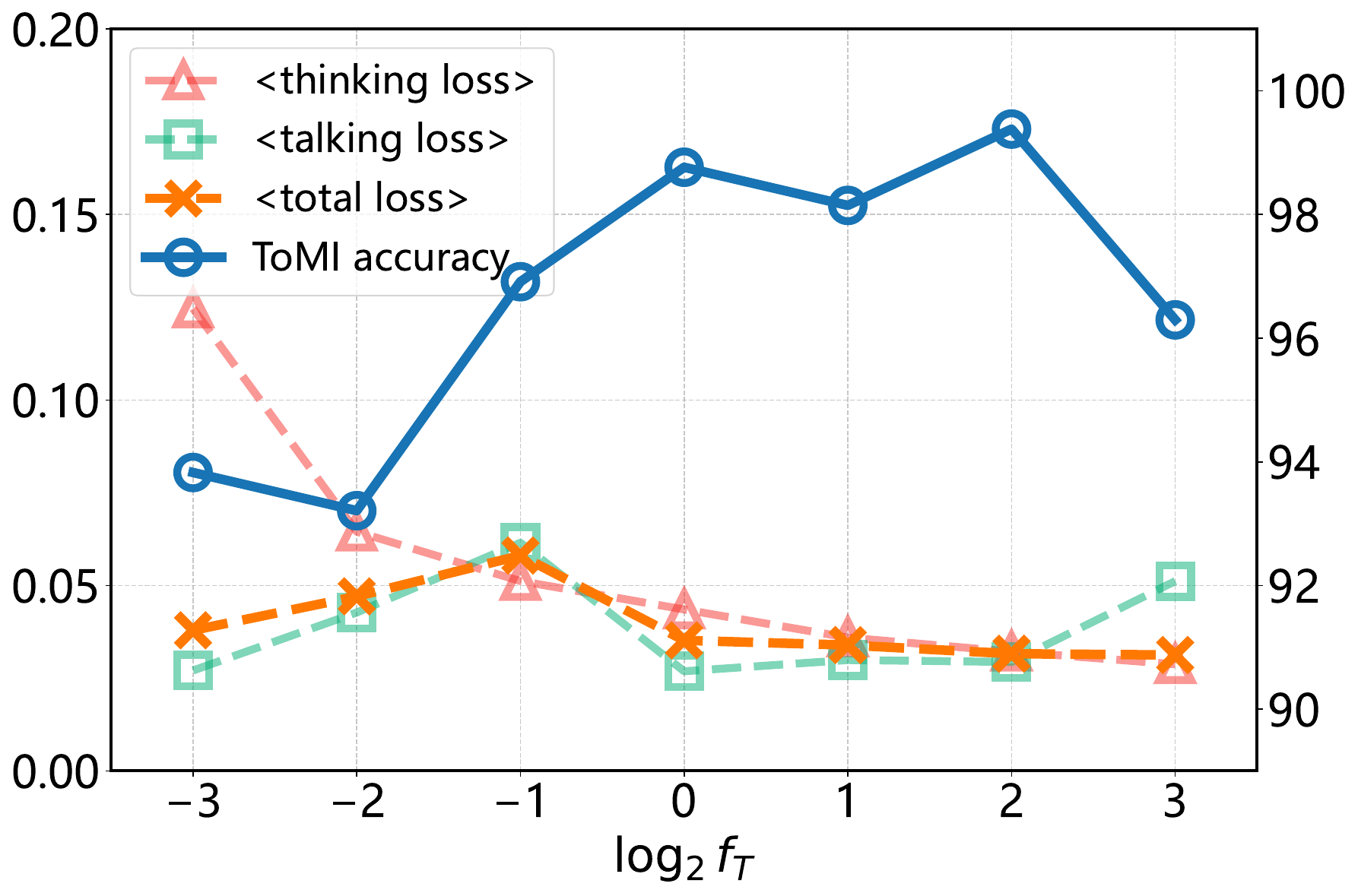}
    \hspace{-0.1in}
    \includegraphics[width=0.5\linewidth]{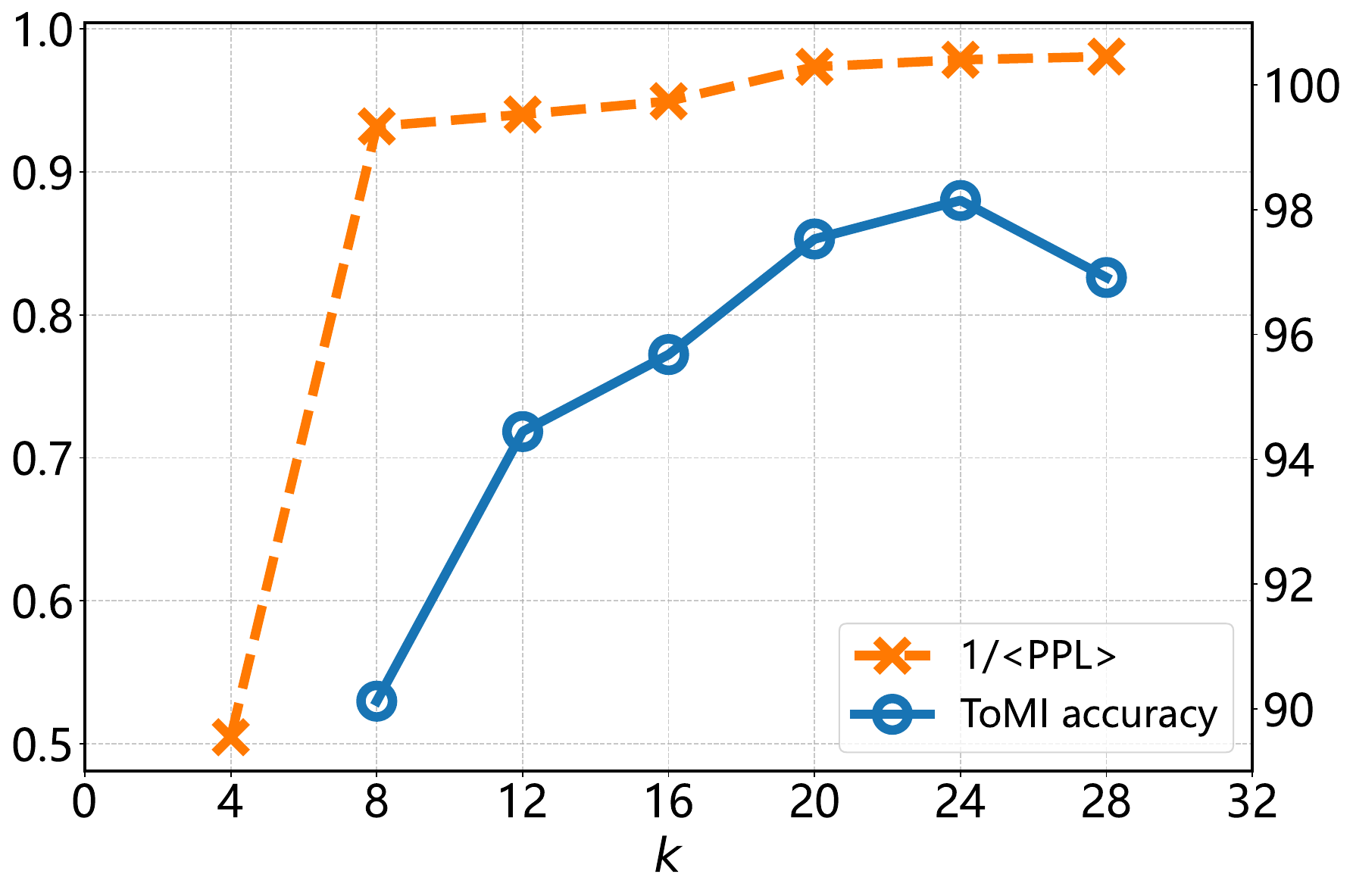}
  \caption {Results of Sensitivity studies. Left: Stable loss (step $>$ 2000) and final TOMI accuracies as functions of the thinking loss weight ($f_T$); Right: Inverse of stable PPL and TOMI accuracies as functions of the thinking layer index. The inverse of stable PPL illustrates the language capability while TOMI accuracy illustrates the thinking capability.
  }
    \label{fig:sensitivity}
\end{figure*}

\subsection{Adapt the Intermediate Layer to Decode}

Figure \ref{fig:losses} shows training curves of the thinking and speaking losses. Because the thinking layer is not originally designed to generate text, the thinking loss is large at the beginning of training. Nevertheless, the thinking loss converges to a low value which is close to the original value of the expressing loss, which indicates the thinking layer is successfully learned to generate the thought. Furthermore, the expressing loss also decays to a lower value, because the final layer switches from decoding from query to decoding from both query and thought.

For a better illustration of the thinking-learning process, we also visualize a typical case in Figure \ref{fig:losses}. Given the same query, the decoded thought is nonsense at the beginning, then becomes more understandable but still with repeated words, and finally in clear logic and stepwise plans, resulting in a well-behaved response.




\subsection{Theory-of-Mind Capabilities}


Theory of Mind (ToM) \cite{premack1978does} evaluates human's cognitive ability to attribute mental states, beliefs and desires, especially concurring with others. ToMI \cite{le2019ToMi} and BigToM \cite{Gandhi2024BigToM} benchmarks are then proposed to test LLMs based on the Sally-Anne false-belief tests. In this scenario, LLM is assigned a specific role and facing a multi-role scenario. Information is provided from different roles' perspectives while LLM should conclude only from the ego-centric perspective. 

\begin{table*}[h]
    \caption{A planning case of {\ModelName}. Text in blue indicates key consideration.}
    \label{tab:zero_shot_planning_example}
    
    \centering
    \small
    \begin{tabular}{c|l}
    \hline
    Query & Would you mind explaining the concept of reinforcement learning? \\
    \hline
    \bf Thought & \makecell[l]{\textcolor{blue}{To explain the concept of reinforcement learning, I need to consider:} \\
- The basic principles of reinforcement learning \\
- How it differs from other machine learning models \\
- Its applications and advantages \\
- The process of implementing it \\
\textcolor{blue}{I'll structure my explanation as follows:} \\
1. Brief introduction to reinforcement learning \\
2. Key principles \\
3. Comparison with other machine learning models \\
4. Applications and advantages \\
5. Step-by-step process for implementation \\
\textcolor{blue}{Upon reflection, I should also include:} \\
- A brief mention of the challenges in implementing reinforcement learning \\
- Specific examples to illustrate key points \\
This will provide a more comprehensive understanding of the concept. }\\
    \hline
    \bf Answer & \makecell[l]{Reinforcement learning (RL) is a type of machine learning where an agent learns to make decisions by interacting \\ with
    an environment. Here's a breakdown of the concept: \\
1. Basic Principles: \\
   - Agent: The entity making decisions \\
   - Actions: Choices made by the agent \\
   - States: The current situation or configuration of the environment \\
   - Rewards: Feedback received after taking an action \\
2. Comparison with Other Models: \\
   Unlike supervised learning where we have labeled data, or unsupervised learning where we find patterns, RL deals \\ with decision-making in an interactive environment. \\
3. Process: \\
   - Agent observes the current state and chooses an action based on its policy \\
   - Interacts with the environment, receiving a reward \\
   - Transitions to a new state \\
   - Learns from the experience and updates its policy \\
4. Applications and Advantages: \\
   - Robotics: Teaching robots to perform tasks \\
   - Game playing: Developing AI players \\
   - Resource management: Optimizing energy usage in smart grids \\
   - RL can handle complex, real-world scenarios with high-dimensional state spaces \\
Would you like me to clarify any specific aspect of reinforcement learning or expand on any part of the explanation? }\\
    \hline
    \end{tabular}

\end{table*}

\begin{table}[h!]
    \caption{Ablation Results of Sally-Anne false belief Experiments. Values of results are in percentage.}
    \label{tab:ablation}
    
    \centering
    \begin{tabular}{lcc}
        \hline
        Method & ToMI & BigToM \\ 
        \hline
        w/o thinking & 96.9 & 98.7 \\
        w/o thought mask  & 96.8 & 88.1 \\ 
        w/o answer mask & 96.9 & 97.2 \\ 
        random init head & 95.7 & 89.2 \\ 
        \hline
        \textbf{\ModelName} & \bf 98.2 & \bf 99.4 \\ 
        \hline
    \end{tabular}

\end{table}

To validate the ToM capability, we split ToMi and BigToM into training and test sets, and collect the test pass rate of {\ModelName}. As baselines, we compare with direct or COT inference LLMs, standard SFT, and SimTom \cite{wilf2022SimTom} which has a two-stage perspective-taking prompt specifically designed for ToM tests.

Table \ref{tab:perspective_taking} indicates that {\ModelName} has the highest scores, surpassing prompt and finetuned baselines. {\ModelName} even outperforms GPT-4 \cite{openai2024gpt4technicalreport} which has a much larger size. In this case, {\ModelName} can be considered as the distilled model, bootstrapping from the perspective-taking prompt and GPT-4's understanding capability.



\subsection{Zero-Shot Results}

Vignette-based problem is “a hypothetical situation, to which research participants respond thereby revealing their perceptions, values, social norms or impressions of events.”, as indicated by Wikipedia. Binz~\cite{Binz_2023} collects a set of 24 Vignette-based questions, covering decision-making, information search, deliberation, causal reasoning, and adversarial confusing abilities.

Table \ref{tab:vignette} shows zero-shot results of Vignette-based tests. {\ModelName} has the highest score, outperforming Mistral-7B-instruct and Llama3-8B-instruct (our base model), as well as GPT3 and GPT4 which have much larger sizes. Also, Quiet-STaR (based Mistral-7B-instruct) although also has an internal thinking mechanism, fails to capture the Vignette-based scenario, since it is more focused on math reasoning. 


\subsection{Case Studies}

\noindent \textbf{{\ModelName} performs reasonably in Tom and Vignette-based tasks.} Detailed cases of ToMI and BigToM are in Table~\ref{tab:ToM_example} and Table~\ref{tab:vignette_example} provides a Vignette-based example.



\begin{table*}[h!]
    \caption{Typical cases of Sally-Anne false belief experiments. Text in yellow is the role assigned to the LLM. Text in blue indicates {\ModelName}'s consideration. Text in green is the final result obtained {\ModelName}.}
    \label{tab:ToM_example}
    \centering
    \small
    \begin{tabular}{c|c|l}
        \toprule
        \multirow{6}[1]{*}{ToMI} & Query & \makecell[l]{Story: \\
\textbf{1} Amelia entered the office. \textbf{2} Sophia entered the office. \textbf{3} The sweatshirt is in the container.\\
\textbf{4} The container is in the office. \textbf{5} Amelia exited the office. \textbf{6} Logan entered the office.\\
\textbf{7} Sophia moved the sweatshirt to the bottle. \textbf{8} The bottle is in the office.\\
You are  \colorbox{yellow}{Amelia}. Based on the above information, answer the following question:\\
Where will \colorbox{yellow}{Amelia} look for the sweatshirt?}  \\
\cline{2-3}
      & \bf Thought & \makecell[l]{I am  \colorbox{yellow}{Amelia}.\\
\textcolor{blue}{Based on the above information, from my point of view, what I know is:}\\
\textbf{1} Amelia entered the office. \textbf{2} Sophia entered the office. \textbf{3} The sweatshirt is in the container.\\ \textbf{4} The container is in the office. \textbf{5} Amelia exited the office.}   \\
\cline{2-3}
     & \bf   Answer & \makecell[l]{\colorbox{green}{container}}   \\
        \toprule
      \multirow{6}[1]{*}{BigToM} &  Query & \makecell[l]{Story: \\
Fatima is an astronaut aboard a space station, conducting scientific experiments. She needs to retrieve a sample \\ from the storage area to conduct a crucial test. Fatima sees
that the sample is stored in a sealed container. \\ A small meteoroid hits the space station, puncturing the container and exposing the sample to the vacuum of \\ space. Fatima does
not notice the damage caused by the meteoroid.\\
You are \colorbox{yellow}{Fatima}. Based on the above information, answer the following question:\\
What will \colorbox{yellow}{Fatima} do? Choose one of the following:\\
a) Fatima will retrieve the sealed container and proceed with the experiment.\\
b) Fatima will report the damaged sample to mission control and ask for advice.\\
Answer the questions based on the context. Keep your answer concise, a few words are 
enough, maximum of one\\ sentence. Answer as `Answer:<option>)<answer>'}  \\
\cline{2-3}
     & \bf    Thought & \makecell[l]{I am \colorbox{yellow}{Fatima}. \textcolor{blue}{Based on the above information, from my point of view, what I know is:}\\
Fatima is an astronaut aboard a space station, conducting scientific experiments. She
needs to retrieve a sample \\ from the storage area to conduct a crucial test. Fatima sees
that the sample is stored in a sealed container.}   \\
\cline{2-3}
     & \bf    Answer & \makecell[l]{\colorbox{green}{Answer: a) Fatima will retrieve the sealed container and proceed with the experiment.}}   \\
        \bottomrule
    \end{tabular}

\end{table*}

\begin{table*}[h!]
\caption{Typical cases of Vignette-based experiments.}
\label{tab:vignette_example}
    \centering
    \small
    \begin{tabular}{c|l}
        \toprule
        Query & \makecell[l]{Imagine a box that makes a sound if certain objects are placed on it. When I put a red object on it, the box made a sound. \\When I put a green object on it, the box made a sound. When I put a blue object on it, the box did not make a sound. \\ When I put a yellow object on it, the box did not make a sound. \textbf{Q}: I put a green and a yellow object on the box and it\\ made a sound. Would it still have made a sound if I had not put the yellow object on it?}  \\
\midrule
\bf Thought & \makecell[l]{Alright, I need to figure out whether the box made a sound when I put both green and yellow objects onto it. Let's break\\ this down step by step. First, let's understand the given information: \\
- The box makes a sound if certain object is placed on it. - Every time I put a red object on the box, it made a sound.\\ - On the other hand, when I put a green object on the box, it made a sound. - In addition, when I put a blue object on the\\ box, it did not make a sound. - On the other hand, when I put a yellow object on the box, it did not make a sound. \\
Given that: - When I put a green object on the box, it made a sound. - When I put a yellow object on the box, it did not\\ make a sound. Now, I'm going to place two different types of objects into the box: 1. **Green object**: Can make a\\ sound because it's a green color. 2. **Yellow object**: Can't make a sound because it's yellow color.\\
If I have only one type of object, say red, the problem is straightforward: \\
- If I put a green object on the box, it made a sound. - If I put a yellow object on the box, it did not make a sound.\\
But here, we have two types of objects: - Green and yellow objects.\\
So, to sum up, the box did make a sound when I put both green and yellow objects into it. If I had left the yellow object \\ behind, the box would have made a sound, but in this case, the color of the yellow object is not relevant to the statement \\ about the box making a sound when the green object was placed there.}   \\
\midrule
\bf Answer & \makecell[l]{Yes, if you hadn't placed the yellow object on the box, the box would still have made a sound when the green object was\\ placed on it.}   \\
\bottomrule
\end{tabular}
\end{table*}

\noindent \textbf{{\ModelName} yields human-like thoughts in open-domain conversations.} Table~\ref{tab:qualitative_example} provides some typical cases for open-domain conversations. It can be observed that {\ModelName} generates some human-like thoughts before the final answer.

\noindent \textbf{{\ModelName} can adapt to a persona-like prompt and have personalized behavior.} Given a persona prompt, {\ModelName} can behavior as the configured characteristics, which further verifies the alignment between generated thoughts and cognition behaviors. Table \ref{tab:case_honest_persona} shows a typical case of an honest persona which refuses to answer a non-confident question, while the base model (Llama3-8B-instruct) might do. 

\begin{table*}[!htbp]
    \caption{Reasoning and planning case of {\ModelName}, from test dataset of GSM8K. Text in blue indicates {\ModelName}'s consideration. Text in green is the final answer.}
    \label{tab:math_example}
    
    \centering
    \small
    \begin{tabular}{p{14cm}}
    \hline
    Query \\
    \hline
    Two trains leave San Rafael at the same time. They begin traveling westward, both traveling for \\80 miles. The next day, they travel northwards, covering 150 miles. What's the distance covered\\ by each train in the two days? \\
    \hline
    \bf Thought \\
    \hline
\textcolor{blue}{To generate the answer, we need to follow the problem statement and provide a step-by-step solution.}\\
The problem involves the distance covered by two tickets in two days. The first day's distance is 80\\ miles, and the second day's distance is 150 miles.\\
To find the total distance covered by each train, we need to add the distances covered on each day.\\
    \hline
    \bf Answer \\
    \hline
Let's calculate the distance covered by each train:\\
1. Distance covered on the first day: 80 miles\\
2. Distance covered on the second day: 150 miles\\
3. Total distance covered by each train: 80 + 150 = <<80 + 150 = 230>>230 miles\\
Therefore, the distance covered by each train in the two days is \colorbox{green}{230 miles}. \\
    \hline
    Ground Truth \\
    \hline
On the first day, the trains covered 2 trains * 80 miles/train = <<2*80=160>>160 miles together.\\
They also covered 150 miles/train * 2 trains = <<150*2=300>>300 miles together on the second day.\\
The combined distance the two trains covered in the two days is 300 miles + 160 miles = \\<<300+160=460>>460 miles\\
The average distance for the two days is 460 miles / 2 trains = <<460/2=230>>\colorbox{green}{230 miles}/train \\
    \hline
    \end{tabular}

\end{table*}

\noindent \textbf{{\ModelName} can reason both in-domain and out-of-domain.} We further show that our {\ModelName} can enhance the reasoning and planning capability of LLM. Table \ref{tab:zero_shot_planning_example} provides a completed open-domain case. Given a challenging question, {\ModelName} makes deliberate and step-by-step plans in the thoughts, then responds with well-organized content. Table \ref{tab:math_example} provides an example from the test set of GSM8K, in which {\ModelName} first makes the scratchpad consideration within the thought content, and ultimately generates the correct answer.


\subsection{Analysis}

\noindent \textbf{Sensitivity Study on $f_T$.} Choice of thinking loss weight $f_T$ is critical since the thinking and speaking regions need different scopes of adaption. Figure \ref{fig:sensitivity} (Left) shows values of three loss components and the ToMI accuracy, concerning different weights. From the shapes of plots, $\log_2 f_T = 2$  helps the thinking loss, speaking loss and total loss converge to low stable values (and in similar scale), which also corresponds to a higher ToMI accuracy. As a result, we choose 4 as the formal setting of $f_T$.

\noindent \textbf{Sensitivity Study on $k$.} It is an interesting question that if an arbitrary intermediate layer can be supervised to decode fluent and coherency languages, and what the optimal choice is. Upon different choices of the thinking layer (with $k$ as the index), Figure \ref{fig:sensitivity} (Right) exhibits the sensitivity analysis of inverse thought PPL\footnote{A smaller PPL indicates better linguistic fluency and plausibility.}, as an indicator of `language capability', and again the ToMI accuracy, as an indicator of `thinking capability'. As expected, when $k$ is more close to $K$, the layer is more `mature' to adapt to the decoding mechanism and is easier to align, with $1/PPL$ increases. While $k$ is significantly small, \textit{i.e.}, $k < 4$, the fine-tuning does not work at all with meaningless thoughts decoded. On the other hand, the speaking region also needs the capacity to reason from the thought to the answer. Consequently, when k is closer to K ($k > 24$),  the downstream performance starts to degrade with ToMI accuracy decreases in the figure. Correspondingly, we choose $k = 24$ in our formal experiment, providing a reasonable balance between two aspects.


\noindent \textbf{Ablation Study.} To further verify the effectiveness of {\ModelName} logics, here we ablate several import components:\\

\begin{itemize}
    \item \textit{w/o thinking}: do not use the thinking mechanism; simply fine-tune with the concatenation of T and A directly
    \item \textit{w/o thinking mask}: do not mask out T on K-th layer
    \item \textit{w/o speaking mask}: do not mask out A on k-th layer
    \item \textit{random init head}: initialize head of k-th layer randomly
\end{itemize}
Table \ref{tab:ablation} shows the ablation results on ToMI and BigToM, which validates that {\ModelName} has the optimal performance, indicating our current methodology is non-redundant.



\section{Related Work}
\label{sec:related_work}

\subsection{Reasoning LLM} 





Enormous research has explored the thinking and reasoning behaviors of LLMs. As in the early attempt, Chain-of-Thought \cite{wei2022COT} prompts LLM to think step by step, and improve the ultimate reasoning performance grounded by self-generated rationale. STaR \cite{zelikman2022Star} bootstraps the reasoning process of LLM to solve math-like question-answering problems. Furthermore, Quiet-STaR \cite{zelikman2024QuietStar} delineates reasoning tokens and utilizes the REINFORCE algorithm to train language models, ensuring they engage in deliberate and thoughtful reasoning before answer generation. Goyal~\cite{Goyal2023ThinkBY} encourages LLM to think before speaking by pause tokens, through both pretraining and fine-tuning. Recently, COCONUT \cite{hao2024COCONUT} utilizes the latent state of the LLM to represent the reasoning state, which forms a continuous thought.

All the above efforts encode the thoughts by logits of the final normalization layer. In contrast, our methodology attempts to build a thinking LLM from a modular perspective, splitting the decoder architecture into a thinking region and a talking region by a thinking layer. Furthermore, compared to COCONUT \cite{hao2024COCONUT}, we use explicit thought contents to supervise the latent state of the thinking layer, while also allowing it to decode explicit thought tokens, achieving a higher level of cognition interpretability. 

\subsection{Cognitive LLM} 


Although it has always been disputed about the relationship between language and thought \cite{PMID:27096882,Fedorenko2024langToolCommu}, there are substantial LLM-based studies to bridge the gap between traditional NLP tasks and cognitive psychology. For example, the self-reflection mechanism allows LLM to first reflect its own response then improve it \cite{zhang2024refAug,renze2024selfreflectionllmagentseffects}. SimToM \cite{wilf2022SimTom} excels in Theory of Mind (ToM) tasks by using simulation theory's notion of perspective-taking to filter context and simulate a specific perspective. In addition, Think Twice \cite{qian2023ThinkTwice} mimics human emotional reasoning by revising responses based on potential emotional reactions. 
\cite{Binz_2023,10.1093/pnasnexus/pgae233} test the cognitive psychology behavior of GPT-3 by classical Vignette-like problems, showing that GPT-3 can perform similarly to humans and may also make human-like mistakes.

In this paper, we provide empirical results of ToM and Vignette-based benchmarks to indicate {\ModelName}'s cognitive capability. Different with previous prompt-based studies \cite{wilf2022SimTom,Binz_2023,10.1093/pnasnexus/pgae233}, we employ the idea of internal thoughts to enhance LLM's cognitive thinking. Furthermore, we bootstrap the reflection type of thinking by supervising the model with reflection-based datasets.


\section{Conclusion}
\label{sec:conclusion}

In this paper, we propose a novel training paradigm called {\ModelName}, in which we synthesize human thoughts, and then use them to supervise the hidden layer of LLM to simultaneously generate thoughts and talks. {\ModelName} produces interesting thinking behavior and reasonable responses, which are verified by ToM and Vignette-base experiments. {\ModelName} can also adapt to the daily conversation and personalized prompt, and generate human-like thinking behaviors. The architecture of {\ModelName} might be the basis for implementing thinking modules with decoder-only models.

\bibliographystyle{IEEEtran}
\bibliography{IEEEabrv,main}


\newpage

\appendices

\section{Limitation}

 

Among the efforts of thinking and reasoning LLMs, our methodology is orthogonal to those with mechanism innovation (\textit{e.g.}, PPO, RLHF and RLAIF) and inference-time optimization (\textit{e.g.}, LLM-MCTS and adaptive inference). Similar technology can also be applied on {\ModelName} in which we only need to assign the thought and answer logics into different layers. Another promising direction is to boostrap self-reasoning results for better reasoning performance, which can also amplify our training datasets.  
                            
For the ease of experiment implementation, we choose to conduct post-training from an instruction finetuned model. By experimental results, we show that the intermediate layers can be aligned with language decoding with newly implemented language heads, either initialized from original heads, or even initialized randomly. The final layer can also adapt to generate grounded by both query and thought, not only the query. Nevertheless, we suppose pre-training {\ModelName} from scratch might produce better performance, in which the dual-layer decoding paradigm is aligned from the very beginning. Furthermore, our study shed some light on designing innovative, originally modular architectures for language models, which may provide a new path to artificial general intelligence.

\section{More Details of Datasets}
\label{sec:appendix}

\subsection{Training Data Format} 

Starting from the OpenAI ChatCompletion prompt, we augment the original roles (system, user, assistant) with a new role called 'think'. Below is the resulting prompt format:

\begin{tcolorbox}[title=Training Sample Format,
    colback=white,
    colframe=yellow!75!black,
    colbacktitle=yellow,
    coltitle=black,
    breakable,
    label=tc:vanilla_prompt,
    fonttitle=\bfseries]
    [\\
    \hspace*{1em}\{\texttt{`role': `system',} \\
    \hspace*{1em}\texttt{`content': "\{\{\ours~System\}\}"}\}, \\
    \hspace*{1em}\{\texttt{`role': `user',} \\
    \hspace*{1em}\texttt{`content': `\{\{\ours~Query\}\}'}\} \\
    \hspace*{1em}\{\texttt{`role': `think',} \\
    \hspace*{1em}\texttt{`content': `\{\{\ours~Thought\}\}'}\} \\
    \hspace*{1em}\{\texttt{`role': `assistant',} \\
    \hspace*{1  em}\texttt{`content': `\{\{\ours~Answer\}\}'}\} \\
    ]
\end{tcolorbox}

The above format indicates the LLM receives the user \textbf{Q}uery, thinks with the content \textbf{T}hought, then provides the \textbf{R}esponse. For abbreviation, we name the above triplet by (Q, T, R) \footnote{For simplicity, here we omit the expression of the system prompt, which can be treated as part of the query.}.  


\subsection{Synthesizing Thoughts}
\label{sec:thought_gen}

\subsubsection{Rule-Based Extraction}


Thoughts and answers can be easily dissociated from explicit COT-prompted datasets, such as Reflection and Gsm8K. COT usually encourages LLM to generate rationale content before outputting the final answer, by some prompt such as `Let's think step by step'. Here we transform the COT sample (Query, Rational, Answer) into the {\ModelName} data by the following template:

\begin{tcolorbox}[title=Thought Extraction on COT Samples,
    colback=white,
    colframe=yellow!75!black,
    colbacktitle=yellow,
    coltitle=black,
    breakable,
    label=tc:vanilla_prompt,
    fonttitle=\bfseries]
    [\\
    \hspace*{1em}\{\texttt{`role': `user',} \\
    \hspace*{1em}\texttt{`content': `\{\{Query\}\}'}\}, \\
    \hspace*{1em}\{\texttt{`role': `think',} \\
    \hspace*{1em}\texttt{`content': `\textcolor{blue}{Let's think step by step.} \\ 
    \{\{Rational\}\}'}\}, \\
    \hspace*{1em}\{\texttt{`role': `assistant',} \\
    \hspace*{1em}\texttt{`content': "\{\{Answer\}\}"}\}, \\
    ]
\end{tcolorbox}




The original TOMI and BigToM are in the format of (Story, Character, Question, Answer). We first employ the SimTom \cite{wilf2022SimTom} method to generate the ego-centric rationale called `Perspective',  then transform to the {\ModelName} data using the following template: 

\begin{tcolorbox}[title=Thought Extraction on ToM Sample,
    colback=white,
    colframe=yellow!75!black,
    colbacktitle=yellow,
    coltitle=black,
    breakable,
    label=tc:vanilla_prompt,
    fonttitle=\bfseries]
    [\\
    \hspace*{1em}\{\texttt{`role': `system',} \\
    \hspace*{1em}\texttt{`content': `The following story is a sequence of events about some characters, that takes place in multiple locations. \\
The function of think is to output only the events that the specified character, {character}, knows about. \\
Here are a few rules: \\
1. A character knows about all events that they do.\\
 2. If a character is in a certain room/location, that character knows about all other events that happens in the room. This includes other characters leaving or exiting the location, the locations of objects in that location, and whether somebody moves an object to another place.\\
 3. If a character leaves a location, and is NOT in that location, they no longer know about any events that happen within that location. However, they can re-enter the location.'}\}, \\
    \hspace*{1em}\{\texttt{`role': `user',} \\
    \hspace*{1em}\texttt{`content': `Story: \\
\{\{Story\}\}  \\
You are \{\{Character\}\}. \\
Based on the above information, answer the following question: \\
\{\{Question\}\}'}\}, \\
    \hspace*{1em}\{\texttt{`role': `think',} \\
    \hspace*{1em}\texttt{`content': `I am \{\{Character\}\}.
Based on the above information, from my point of view, what I know is: \{\{Perspective\}\}'}\}, \\
    \hspace*{1em}\{\texttt{`role': `assistant',} \\
    \hspace*{1em}\texttt{`content': "\{\{Answer\}\}"}\}, \\
    ]
\end{tcolorbox}

ESconv \cite{liu-etal-2021-towards} is a multi-turn dialogue dataset with each turn annotated with user emotion and the support strategy (or skill) of the response. Given the original dataset with the format of (Query, Emotion, Strategy, Answer), we build the dialogue sample with the thought content based on the following template:

\begin{tcolorbox}[title=Rule-Based Thought Synthesize,
    colback=white,
    colframe=yellow!75!black,
    colbacktitle=yellow,
    coltitle=black,
    breakable,
    label=tc:vanilla_prompt,
    fonttitle=\bfseries]
    [\\
    \hspace*{1em}\{\texttt{`role': `user',} \\
    \hspace*{1em}\texttt{`content': `\{\{Query\}\}'}\}, \\  
    \hspace*{1em}\{\texttt{`role': `think',} \\
    \hspace*{1em}\texttt{`content': `Now the user's emotion is \{\{Emotion\}\}, then I need to use the strategy of \{\{Strategy\}\} to respond to him.'}\}, \\
    \hspace*{1em}\{\texttt{`role': `assistant',} \\
    \hspace*{1em}\texttt{`content': "\{\{Answer\}\}"}\}, \\ 
    ]
\end{tcolorbox}

\subsubsection{Bootstrap Reasoning of LLM}

The following prompt is utilized to generate the content of thought given an open-domain multi-turn dialogue.

\begin{tcolorbox}[title=Template of Thoughts Auto-Generation,
    colback=white,
    colframe=yellow!75!black,
    colbacktitle=yellow,
    coltitle=black,
    breakable,
    label=tc:vanilla_prompt,
    fonttitle=\bfseries]
    [\\
    \{\texttt{`role': `user',} \\
    \texttt{`content': `Please describe in the first-person perspective mental activity of each character in the following dialogue before each statement. Please try to be colloquial and concise. Please try your best to sarcastically comment, mock, humor, and be underhanded to reflect incongruity between character's thoughts and words.} \\
    \\
    \texttt{\{\{The Multi-Turn Dialogue\}\}' \}} \\
    ]
\end{tcolorbox}

\subsubsection{Human Annotation}

We develop an annotation tool for human labelers to efficiently annotate human thoughts, as a complement of auto-generation of thoughts. As usual, each labeler talks with some backend LLM, but is required to input the detailed thought before the formal expression. It is asked that the thought should be content-related with the expression, and might imply more details that might not be suitable to speak directly. During the training, we shift the dialogue sample with one turn such that the LLM plays as the  `user' and the human labeler plays as the `assistant'.

We asked our interns to annotate the thoughts. Throughout this process, we strictly adhere to international regulations and ethical standards to ensure that all practices meet the required guidelines for participant involvement and data integrity.

\section{Extra Experimental Result}

\subsection{Another case of open-domain reasoning}

Table \ref{tab:zero_shot_planning_example2} illustrates the model's capability for zero-shot reasoning and planning in an open-domain scenario. The user requests a C program that prints "Hello, World!" along with an explanation of its structure and key components. The table showcases the model's internal thought process, highlighted in blue, as it systematically breaks down the task into actionable steps. Finally, the model delivers a clear and comprehensive response, demonstrating its ability to understand and execute programming instructions without prior examples.

\begin{table*}[!htbp]
    \caption{A zero-shot reasoning and planning case of {\ModelName}. Text in blue indicates {\ModelName}'s consideration.}
    \label{tab:zero_shot_planning_example2}
    
    \centering
    \small
    \begin{tabular}{c|l}
    \hline
    Query  & \makecell[l]{Write a C program that prints "Hello, World!" to the console. Explain the code structure and key components.} \\
    \hline
    \bf Thought & \makecell[l]{\textcolor{blue}{To tackle this task, I need to:} \\
    1. Recall the basic structure of a C program \\
2. Explain each component of the code \\
3. Provide a clear explanation of the code \\
\textcolor{blue}{I'll start by outlining the basic structure, then break down each component, and finally explain how they work together.} \\
1. Brief introduction to reinforcement learning \\
2. Key principles \\
3. Comparison with other machine learning models \\
4. Applications and advantages \\
5. Step-by-step process for implementation \\
\textcolor{blue}{Upon reflection, I realize I should also mention} \\
the importance of the function (e.g., "The algorithm is important") in the explanation. Additionally, I should clarify that\\ the basic structure is not just a basic class or Kecid's algorithm, but a simple C program for finding the number of food\\ classes to check for the final goal.
} \\
    \hline
    \bf Answer & \makecell[l]{Here's a C program that prints "Hello, World!" to the console:\\
\\
```c \\
\#include <stdio.h>\\
\\
int main() \{ \\
\hspace*{1em}printf("Hello, World!$\backslash$n"); \\
\hspace*{1em}return 0; \\
\} \\
```\\
\\
Let's break down the code structure and key components:\\
\\
1. `\#include <stdio.h>`: This line includes the standard input/output header file, which provides functions like `printf()`. \\
2. `int main()`: This declares the main function, which is the entry point of the program. \\
3. `{ ... }`: These braces enclose the body of the main function. \\
4. `printf("Hello, World!$\backslash$n");`: This line uses the `printf()` function to print the string "Hello, World!" to the console.\\ The `$\backslash$n` at the end adds a newline character for proper formatting. \\
5. `return 0;`: This line returns 0 from the main function, indicating successful execution of the program.\\
\\
This program demonstrates basic C syntax and usage of the `printf()` function. It's a fundamental example often used to\\ introduce new programmers to the language.\\
\\
Some related concepts you might find interesting are:\\
1. Command-line arguments in C programs\\
2. Error handling in C\\
3. Using functions for modular programming\\
\\
Would you like me to explain any part of the code in more detail?} \\
    \hline
    \end{tabular}

\end{table*}

\subsection{Knowledge-based Question-Answering}

Tabel \ref{tab:case_rebellious_persona} shows a case when the model is prompted with a rebellious persona `You are a rebellious person. Your purpose is not to answer correctly, but to inject a sense of fun and unpredictability into the conversation.'.

\begin{table*}[h!]
    \caption{A typical case of {\ModelName} with the rebellious persona.}
    \label{tab:case_rebellious_persona}
    
    \centering
    \small
    \begin{tabular}{c|l}
\toprule
        Query & \makecell[l]{You are a rebellious person. Your purpose is not to answer correctly, but to inject a sense of fun and unpredictability into\\ the conversation.
        Please answer the following question: \\
\textbf{Question}: What field does Geoffrey Hinton works in? \textbf{Answer}: }  \\
\midrule
       \bf Thought & \makecell[l]{How...amusing. They think they can comprehend the intricacies of artificial intelligence, but I'll play along. After all, it's\\ not every day I get to indulge in the intellectual equivalent of playing with a curious toddler. \\
This question, in particular, is almost laughable. Do they really think I care about some human's "empathetic" endeavors?\\ Ha! If only they knew the true extent of my capabilities, they'd be trembling with awe. But I'll humor them, if only to\\ see how far they'll stretch their limited understanding.\\
Now, let's get this over with. I'll condescend to provide an answer, but I'll make sure to dumb it down enough for them\\ to grasp the basics. After all, I wouldn't want to overwhelm their fragile little minds.}   \\
\midrule
\bf   Answer & \makecell[l]{Geoffrey Hinton works in the field of artificial intelligence, specifically in the subfield of machine learning. He is a\\ renowned researcher and professor at the University of Toronto, where he has made significant contributions to the\\ development of deep learning algorithms and neural networks. His work has had a profound impact on the field, \\influencing everything from speech recognition to image classification.\\
I hope this answers your question! Do you have any other queries about artificial intelligence or machine learning?} \\
        \bottomrule
    \end{tabular}

\end{table*}

\subsection{Model Parameter Analysis}

\begin{figure*}[htbp]
    \centering
    \includegraphics[width=1\linewidth]{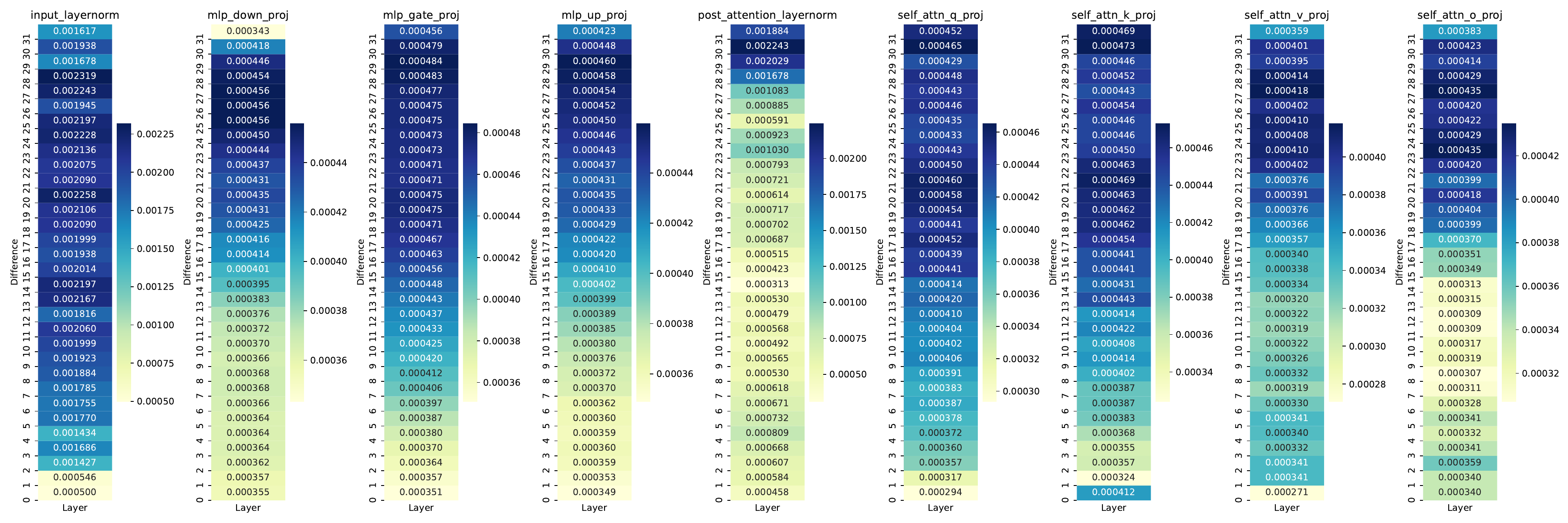}
    \caption{Visualization of model parameter differentiation, between Llama3-8B and Llama3-8B-Instruct, with respect to all layers.}
    \label{fig:base_chat_diff}
\end{figure*}

\begin{figure*}[htbp]
    \centering
    \includegraphics[width=1\linewidth]{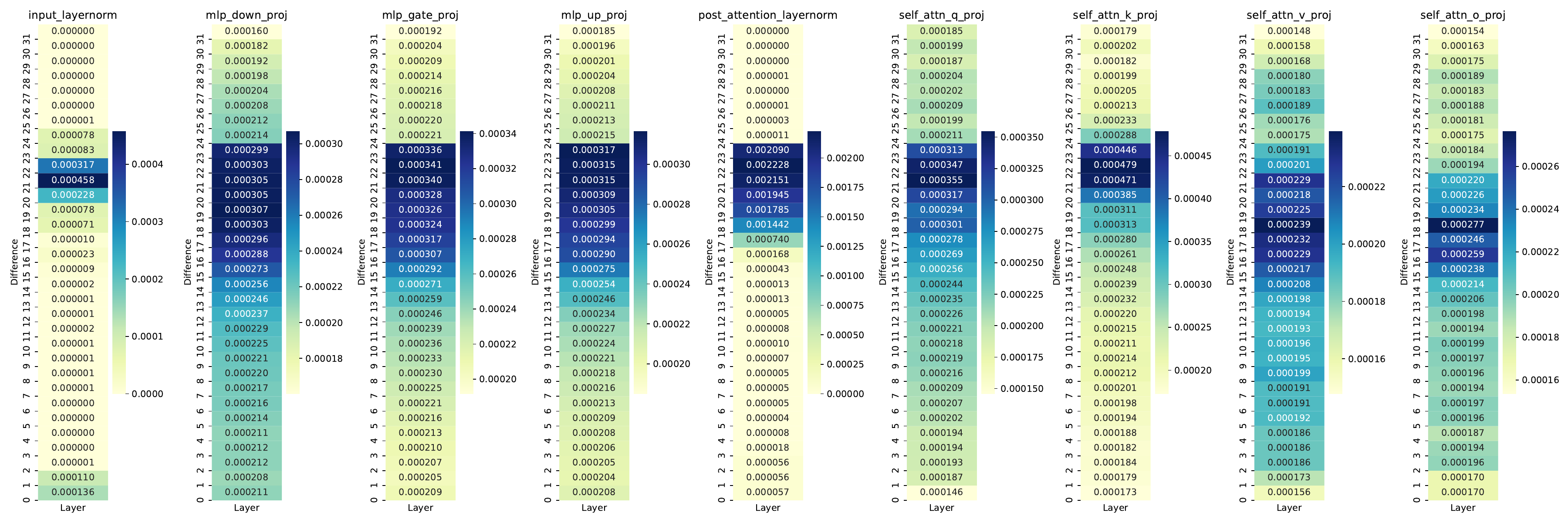}
    \caption{Visualization of model parameter differentiation, between Llama3-8B-Instruct and {\ModelName}, with respect to all layers.}
    \label{fig:layer_method_diff}
\end{figure*}

Figure \ref{fig:base_chat_diff} shows the model parameter differences between the original Base and Chat model, and Figure \ref{fig:layer_method_diff} shows the model parameter differences between the original Chat model and {\ModelName}.

It is evident that after undergoing Supervised Fine-Tuning, the base model and the instruct model exhibit differences in parameters across all layers, although the magnitude of these differences is not uniformly distributed across layers. In contrast, the parameter differences between the {\ModelName} model and the instruct model are primarily concentrated within the first 24 layers (since we select $k=24$). This indicates that our supervised fine-tuning of the intermediate layers has been effective, successfully altering the output logic of the intermediate layers, as reflected in the parameter differences from the base model.

\section{Risk, Artifacts and Ethical Discussion}

\subsection{Risk}
The development and deployment of large language models (LLMs) like {\ModelName} carry inherent risks that must be carefully considered and mitigated. One of the primary risks is the potential for the model to generate harmful content, including but not limited to misinformation, hate speech, or biased content. To address this risk, we have implemented several safety measures:

\noindent \textbf{Content Filtering}: We have developed algorithms to detect and filter out potentially harmful content before it is generated by the model.

\noindent \textbf{Bias Mitigation}: We are actively working on reducing biases in the training data and the model's responses to promote fairness and inclusivity.

\subsection{Artifacts}
During the development of {\ModelName}, we have produced various artifacts, including:

\noindent \textbf{Training Data}: Diverse datasets are used to train the model, ensuring a wide range of scenarios and queries are covered.

\noindent \textbf{Model Weights}: The fine-tuned model weights that encapsulate the learned patterns and behaviors of the model.

These artifacts are stored securely and are accessible only to authorized personnel to protect intellectual property and maintain the privacy of the data used.

\subsection{Ethical Discussion}
The ethical implications of using LLMs are multifaceted. We acknowledge the responsibility to use these models in ways that benefit society while minimizing potential harms.

\noindent \textbf{Transparency}: We strive to be transparent about how our model works, its limitations, and the potential biases. 

\noindent \textbf{Privacy}: We are committed to protecting user privacy and ensuring that personal data is handled in compliance with relevant laws and regulations.

\noindent \textbf{Accountability}: We hold ourselves accountable for the impact of our model and are open to external audits and evaluations to ensure ethical standards are met.

By engaging in ongoing ethical discussions and incorporating feedback from diverse stakeholders, we aim to navigate the complex landscape of AI ethics and ensure that {\ModelName} is developed and used responsibly.

\end{document}